\documentclass[10pt,twocolumn,letterpaper]{article}

\usepackage{times}
\usepackage{epsfig}
\usepackage{graphicx}
\usepackage{amssymb}
\usepackage{float}
\usepackage{balance}
\usepackage{amssymb}% http://ctan.org/pkg/amssymb
\usepackage{pifont}% http://ctan.org/pkg/pifont
\newcommand{\cmark}{\ding{51}}%
\newcommand{\xmark}{\ding{55}}%

\usepackage[breaklinks=true,bookmarks=false]{hyperref}

\begin{document}

%%%%%%%%% TITLE
\title{Learning Higher-order Object Interactions for Keypoint-based Video Understanding}

\author{Yi Huang\thanks{Work done as a NEC Labs Intern}, Asim Kadav, Farley Lai, Deep Patel, Hans Peter Graf\\
NEC Labs America\\
%Princeton, NJ\\
{\tt  {yihuang, asim, farleylai, dpatel, hpg@nec-labs.com}}
% For a paper whose authors are all at the same institution,
% omit the following lines up until the closing ``}''.
% Additional authors and addresses can be added with ``\and'',
% just like the second author.
% To save space, use either the email address or home page, not both
}

\maketitle
% Remove page # from the first page of camera-ready.
% \ificcvfinal\thispagestyle{empty}\fi

%%%%%%%%% ABSTRACT
\begin{abstract}
    % Problem
    Action recognition is an important problem that requires identifying actions in video by learning complex interactions across scene actors and objects. However, modern deep-learning based networks often require significant computation, and may capture scene context using various modalities that further increases compute costs. Efficient methods such as those used for AR/VR often only use human-keypoint information but suffer from a loss of scene context that hurts accuracy.
  
    % Solutions
    In this paper, we describe an action-localization method, KeyNet, that uses only the keypoint data for tracking and action recognition. Specifically, KeyNet introduces the use of {\em object based keypoint information} to capture context in the scene.
    Our method illustrates how to build a structured intermediate representation that allows modeling
    higher-order interactions in the scene from object and human keypoints without using any RGB information. We find that KeyNet is able to track and classify human actions at just 5 FPS. More importantly, we demonstrate that object keypoints can be modeled to recover any loss in context from using keypoint information over AVA action and Kinetics datasets.
\end{abstract}

%%%%%%%%% BODY TEXT
\section{Introduction}
\label{sec:intro}

\begin{figure}[t]
    \centerline{\includegraphics[width=1.0\columnwidth]{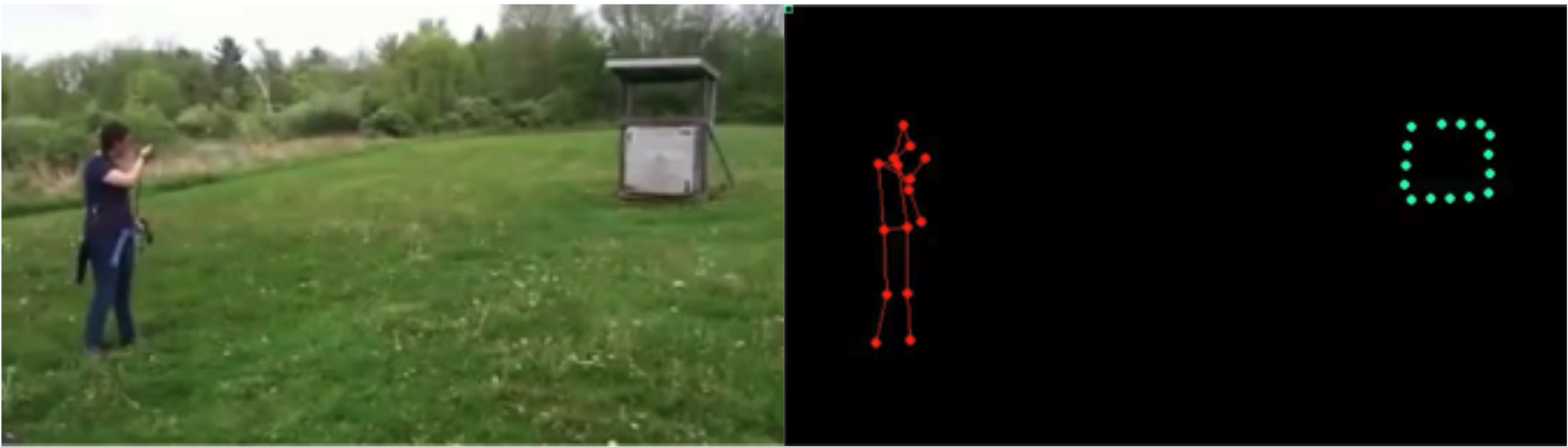}}
    \caption{Left: RGB based action recognition Right: Proposed Human and object keypoint based action recognition}
    \label{fig:fig1}
\end{figure}

% video understanding
Video understanding tasks such as action recognition have shown tremendous progress in the recent years towards various applications in video organization, behavioral analyses etc. Recent methods use parameter intensive 3D convolution or transformer-based networks over RGB data to reach state of the art results\cite{Girdhar2019,Snower2019,Yan2018,Ma2017,Pan2020,Feng2021,Shao2020, Ma2019,Gavrilyuk2020}. These methods are often expensive due to the large amount of computation involved in processing videos. However, many applications especially those in AR/VR often function in resource-constrained settings and often limit to using keypoint based human pose information that can be captured efficiently using hardware sensors. A major draw back of keypoint based methods is that they miss contextual information reducing the overall accuracy.

To address this problem, we develop a method that uses object and human keypoints to understand videos. By integrating object keypoints in our action recognition pipeline, we can recover the scene context information that is lost by using human keypoints. We propose capturing object keypoint information using the Pavlidis\cite{pavlidis1982} algorithm over an existing real-time segmentation method\cite{He2020}. This information can also be alternatively obtained from hardware sensors such as WiFi or radar based sensors~\cite{wang2016interacting, wang2019can}. This generates significant keypoint data over multiple frames that can be difficult to learn. Hence, we structure the joint and keypoint information in intermediate space using a transformer based architecture with joint and positional embeddings that allow KeyNet to recover contextual information and train for the action recognition and localization tasks.

% Problem for previous methods
%Recently, RGB-based action recognition methods had a significant performance boost due to the success of the I3D network and
%keypoints-based methods also leverage the advance of the graph-convolution network(GCN) to achieve more accurate performance.
%However, we argue that it is challenging to deploy these methods for industrial application. RGB-based methods are extremely computationally expensive,
%since it takes dense RGB image sequence as input and utilizes a series of 3D convolution layers to extract features.
%Also,  keypoint-based methods sacrifice important context features and are not able to sense the interaction between actors and various objects.
%Therefore, as constructed in figure \ref{fig:fig1}, we proposed a solution that converts video into a structure representation including both human and object keypoints.

% Our contribution
In this setting, our method is not only is capable of preserving the advantage of the computation efficiency of keypoints-based methods
but is also able to compensate for the loss of context information with our proposed context-aware structure representation.
The primary contributions of our work can be summarized as three aspects:
1) We propose a context-aware structure representation using human and object keypoints in videos. To the best of our knowledge,
it is the first work that utilizes the sub-sampled keypoints to provide context features of objects.
2) We propose KeyNet, a transformer-based network that successfully model the higher-order interaction of various actors and object in videos.
3) On various datasets, we demonstrate that our KeyNet architecture achieves superior performance as compared to the prior convolution-based methods and is an efficient video understanding method for real-world applications.

\section{Related Work}
We discuss and compare against other video understanding methods.

\vspace{2mm}
\textbf{RGB and multi-modal video understanding}
Recent work on action recognition often use 2D/3D convolutions, optical flow and transformer based methods to learn relationships over spatio-temporal elements. 
For example, a large body of existing work uses the output from convolution blocks and aggregates the intermediate features. This representation is then pooled, along with LSTM or other building blocks to learn the temporal information. In contrast, 3D convolution methods, learn the temporal information with the spatial information. For example, some proposed methods, \cite{Lin2019, Pan2020, Feng2021, Shao2020} use a short video snippet as an input and use a series of deep convolution networks to capture the spatial-temporal features. Other methods such as I3D networks \cite{Carreira2017}, and their generated features have been used for variety of video understanding tasks. For example, SlowFast networks combines the knowledge between fast frame rate and slow frame rate video to obtain high accuracy\cite{Feichtenhofer2019}.
% methods based on two-stream
Multi-stream-based methods\cite{Feichtenhofer2012, Wang2016, Wu2016, Carreira2017, Ma2019, Gavrilyuk2020} combine information from video frames and other modality, such as optical flow, human pose, and audio.
They use multiple streams of deep convolution networks to model the knowledge from different modalities and leverage fusion techniques\cite{Feichtenhofer2016} to integrate the knowledge for action recognition. There are several methods that capture human-object interactions in the RGB space, often explicitly using the object information in the scene by using an object detector or convolutional feature maps to capture extract objects in the scene~\cite{gkioxari2018detecting, Ma2017}.

% % Our method
% Distinct from these methods, we explore human action recognition task based on only keypoints to build a video structure representation that is aware of both humans and objects
\vspace{2mm}
\textbf{Keypoint-based Methods}
Existing work over Keypoint-based action recognition uses the skeleton-based action recognition. Existing work follows the classification by detection approach or a top-down approach. Here, the first step is to estimate the keypoints and then use this information to create ``video tracklets'' of human skeletons, learning classification or localization tasks over this intermediate representation. For example,
ST-GCN\cite{Yan2018} uses graph convolution networks to jointly learn the relation of each human body part across each actor.
Other work \cite{Liu2020, Obinata2021} extend this work with addition edges to reasonably aggregate the spatial temporal information in videos. Early work in this area follows the RGB methods, extracting the pose features and then using RNN/LSTMs to learn the temporal information~\cite{du2015hierarchical, du2017rpan, song2017end}. These methods do not capture any object information, and often limit to basic human pose-based action classes such as ``walking'', ``dancing'' etc. Another work, captures the object interactions but uses a separate RGB stream to learn objects and fuses it using a relational network~\cite{wang2018pose}.

\vspace{2mm}
\textbf{Our Work.}
Our work intends to design a context-aware structure representation for videos that is aware of not only actors but also the interactive objects.
Distinct from Non-keypoints-based action recognition, our method only uses sparse information in videos as input and models the knowledge with a lightweight model, therefore, make it more computationally efficient.
Different from the skeleton-based methods, we build our structure representation by using the human and objects keypoints, early in our video representations. This allows the network to learn human-object interaction in the keypoint space, but introduces additional complexity in the intermediate space, where a large amount of keypoint information is introduced. In the next section, we introduce, how we structure this intermediate space to allow transformer networks learn from this information.
\section{KeyNet}

\begin{figure*}[!t]
    \centerline{\includegraphics[width=1.5\columnwidth]{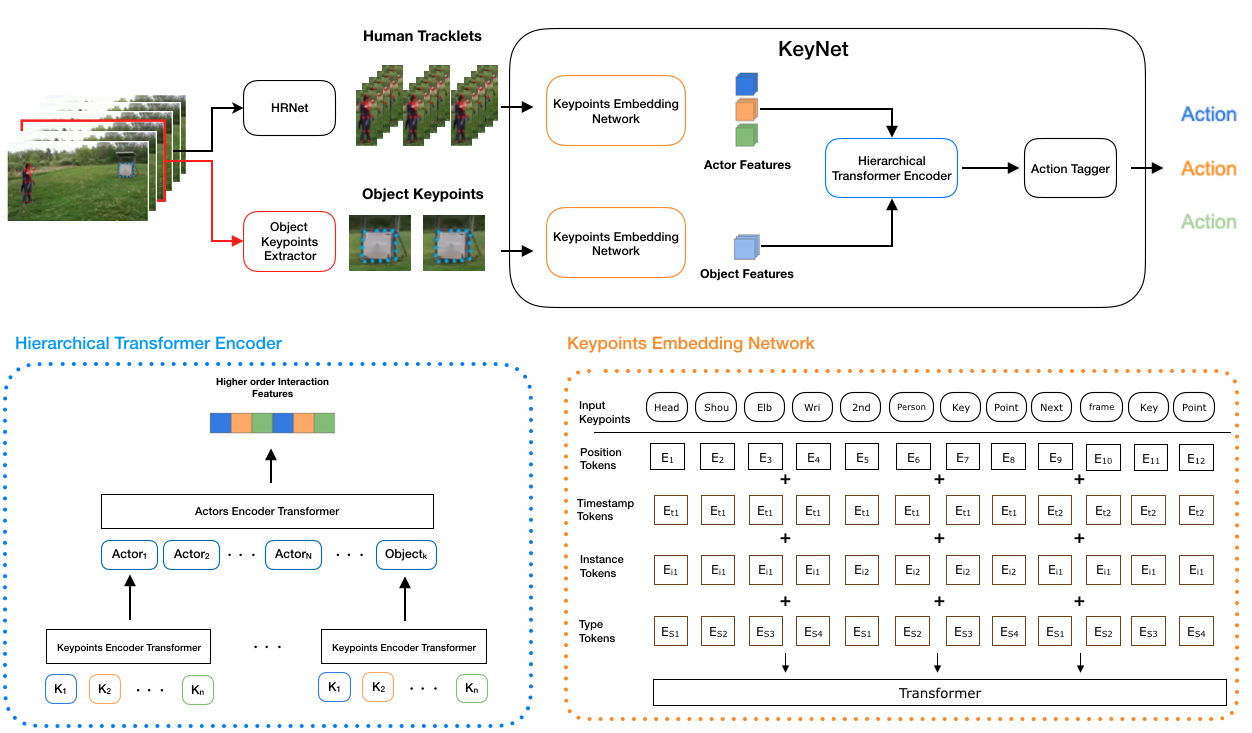}}
    \caption{Flowchart for our proposed KeyNet.}
    \label{fig:KeyNet}
\end{figure*}

In this section, we describe the overall design of our proposed Keynet as shown in Figure \ref{fig:KeyNet}.
Our primary goal is to validate the hypothesis that using sparse keypoints can generate a representation that is sufficient to learn the interactions
between each actor and the background context information.

% cite tubelet action recognition
The model consists of three stages and establishes a tubelet based action recognition pipeline. First, we estimate a set of human and object keypoints for $T$ frames video clip.
Second, the Keypoints Embedding Network projects the keypoints to more representative features by introducing positional embeddings that introduces position, segment and temporal information to the keypoints. Finally, an Action Tagger Network learns the higher-order interactive features and assigns action tags for each actor or predict the action label for the video, depending on the dataset.
We introduce the proposed Action Representation in Section \ref{subsec:action_represnetation}, the Keypoints Embedding Network in Section \ref{subsec:kps_embed},
and the Action Tagger Network is described in Section \ref{subsec:het_net}.

\subsection{Action Representation}
\label{subsec:action_represnetation}

%% assumption
\textbf{Scene Sequence.} We designed the keypoints-based action representation in KeyNet as a scene sequence $D$ where $H_i$ denotes the set of $k_h$ keypoints in the \(i_{th}\) human tracklets and $O_j$ denotes the set of $k_o$ keypoints from the \(j_{th}\) objects.
\[ D = (H_1, H_2 ...H_N, O_1, O_2 ... ,O_K) \]
\[H_i = (P_1, P_2, ... ,P_{k_h})\] 
\[O_j = (P_1, P_2, ... P_{k_o})\]

To obtain a scene sequence $D$ for action representation, we proposed a keypoints sampling method to extract $N$ human tracklets as $H_i$ for actor features
and $M$ objects keypoints as $O_j$ for contextual features.

\begin{figure}[t]
    \centerline{\includegraphics[width=0.8\columnwidth]{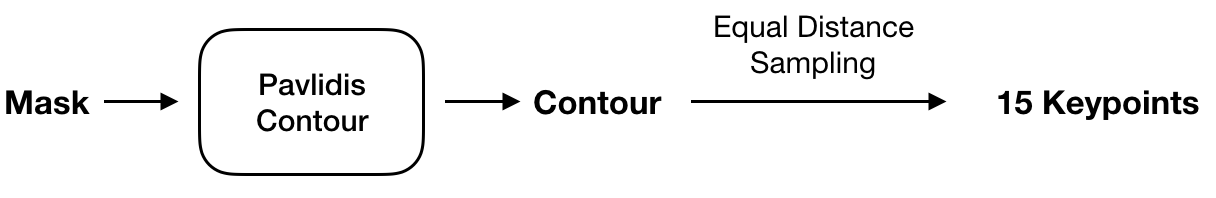}}
    \caption{We extract object keypoints over masks using Pavidilis algorithm}
    \label{fig:flow_obj_kps}
\end{figure}

The object keypoints are introduced to compensate the context information loss in the scene information, often observed in keypoints-based methods.
\vspace{2mm}
\newline
%% Human Keypoints Extractor
% HRNet + Traction
\textbf{Human Tracklet.} To get $N$ human tracklets, we combine a person detector with simple IOU-based tracker, to build a person tubelets over $T$ frames.
Then, we use the HRNet keypoints estimator is used to extract $P$ human joints information for each detected person over $T$ frames~\cite{Wang2019}.
More precisely, for our person detector, we follow previous works \cite{Feichtenhofer2019} to apply Faster R-CNN with ResNeXt-101-FPN backbone.
This detector is pretrained on COCO and fine-tune on AVA with mAP 93.9AP@50 on the AVA validation set.
Regarding keypoints Estimator, we use HRNet\cite{Wang2019} pretrained on PoseTrack with 81.6\% AP on PoseTrack18 validation set.
By selecting the top $N$ person based on the detection confidence score, we can form a human tracklet $S$ sequence with $N * P * T$ keypoints.
\vspace{2mm}
\newline
%% Object Keypoints Extractor
\textbf{Object Keypoint.} We extract object keypoints is to provide contextual features in scenes to enhance
the performance for those object interactive actions. We proposed that human-object interactive action can be modeled by a set of class-agnostic keypoints
with only the shape and spatial information about the object.
Therefore, we extract the object keypoints by performing a sub-sampling along the contour of the mask detected by Mask R-CNN\cite{He2020}.
The flowchart for extracting keypoints is shown in Figure \ref{fig:flow_obj_kps}

% mask rcnn + pavlids + equal distance sampling
More specifically, for each video clip, we apply Mask R-CNN on its keyframe to collect the class-agnostic masks and for each object mask.
For contour tracing, we leverage the Theo Pavlidis' Algorithm \cite{pavlidis1982} to obtain a set of keypoints around each detected object.
Finally, by applying an equal distance sampling on the contour, we extract the keypoints that have the same interval along the contour of the detected mask.
Hence, by selecting the top $M$ object with the highest confidence scores, we can obtain $B$ with $K * P$ keypoints for each  $T$ frames video clips.

\subsection{Keypoints Embedding Network}
\label{subsec:kps_embed}

In this section, we describe how to build an intermediate structured information using no RGB data, and with just person and object keypoints to perform action classification.
To effectively learn the knowledge of action in keypoints representation, we need the information of the spatial correlation between each joint as well as
how these joints may evolve through time. Therefore, we embed this information into the scene sequence by first converting each keypoint in a scene sequence to a sequence of Token
and linearly projecting each Token into an embedding $E$, a learnable lookup table to model the relationship of each keypoint.

\vspace{2mm}
\textbf{Tokenization:} The goal of tokenization is to address extra spatial temporal information and convert it into a more representative information for learning the interaction between keypoints.
To achieve this goal, we extend the prior tokenization techniques \cite{Snower2019} by adding an additional instance token in the  embedding representation for our experiments.
For Position Token, Type Token and Segment Token, we follow previous work \cite{Snower2019} to provide each keypoints with representations of spatial location, temporal location index, and the unique body type information (e.g. Head, Shoulder, and Wrist.) respectively.
Our addition of extending the Segment Token to $T$ time frames and addressing the idea of Instance Token to indicate the id of tracklets that keypoints belong to in the current scene allow the network to learn localization information in the scene.
We generalize the application of previous tokenization methods from pair-wise matching to jointly provide information of the spatial-temporal correlation of multiple instances at the same time.
For the equation below, we described how to convert a scene sequence to 4 types of tokens:

\begin{figure}[h]
    \centerline{\includegraphics[width=1.0\columnwidth]{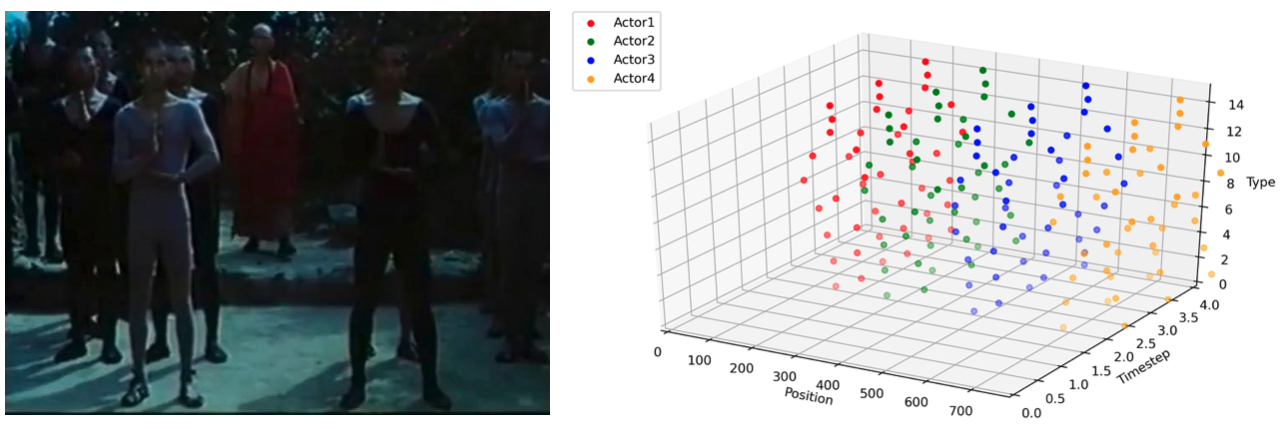}}
    \caption{The visualization for our proposed Position Token, Type Token, Segment Token and Instance Token. The x, y and z axis represents the value of Position Token, Segment Token and Type token respectively and the color denotes the value of Instance Token.}
    \label{fig:token}
\end{figure}

\vspace{2mm}
\textbf{Position Token}\cite{Snower2019}: The down-sampled spatial location of original image and gives the unique representation of each pixel coordinate.
For a keypoints $P$, we write its Position Token as $\rho$, whose range lies in $[1, W']$, $[1, H']$. It reduce the computation cost while preserving the spatial correlation of each keypoints in image.
The general expression of Position Token is below, where $\rho^{p^{t}_k}_n$ indicates the Position Token of the $k_{th}$ keypoint for the $n_{th}$ person in timestamp $t$.

\begin{equation}
    \label{eq:position_token}
    \{ \rho^{p^{t}_1}_1, \rho^{p^{t}_2}_1 \cdots \rho^{p^{t}_1}_2, \rho^{p^{t}_2}_2 ... \rho^{p^{t-T}_K}_{N-1} \cdots \rho^{p^{t-T}_K}_N \}
\end{equation}

\vspace{2mm}
\textbf{Type Token}\cite{Snower2019} The Type Token represents the characteristic of human body(i.e. Head, Right Shoulder and Left Wrist). It is range from $[1, K]$ where $K$ is the number of keypoints. It provides the knowledge of how each part of human body evolves in the keypoint sequence, which is an essential to achieve high accuracy at low resolutions. We assign the Type Token of a keypoint  $P^{p^{t}_k}_n$ as $k$ and the Type Token for the $n_{th}$ person in timestamp $t$ can be written as $k^{p^{t}}_n$ . A general expression for Type Token is shown below

\begin{equation}
    \label{eq:Type Token}
    \{ 1^{p^{t}}_1, 2^{p^{t}}_1 \cdots 1^{p^{t}}_2, 2^{p^{t}}_2 \cdots (K-1)^{p^{t-T}}_N \cdots K^{p^{t-T}}_N \}
\end{equation}

\vspace{2mm}
\textbf{Segment Token} The Segment token provides the difference between the timestamp of keypoints $p^t$ and the timestamp of key-frames. In our modelling of the video scene sequence, the range of Segment token is from [1, T] where T is the total number of frames in a video clip. We assign the Segment Token of a keypoint  $P^{p^{t}_k}_n$ as $t$ and the Segment Token for the $k_{th}$ keypoint from the $n_{th}$ person can be written as $t^{p_{k}}_n$.
The general expression of the Segment token is shown in Equation \ref{eq:Segment Token}

\begin{equation}
    \label{eq:Segment Token}
    \{ 1^{p^{t}_1}_1, 1^{p^{t}_2}_1 \cdots 1^{p^{t}_1}_2, 1^{p^{t}_2}_2 \cdots T^{p^{t-T}_K}_{N-1} \cdots T^{p^{t-T}_K}_N \}
\end{equation}

\vspace{2mm}
\textbf{Instance Token} The Instance Token provides the spatial correlation for a keypoint $P^t$ that provides instance correlation within a frame. It serves a similar role as the Segment Token, providing spatial instead of temporal information. We assign the Instance Token of a keypoint  $P^{p^{t}_k}_n$ as $n$ and the Instance Token for the $k_{th}$ keypoint in timestamp $t$ can be written as $n^{p^t_{k}}$.
The general expression of the Segment token is shown in Equation \ref{eq:inst_token}

\begin{equation}
    \label{eq:inst_token}
    %  correct the equation
    \{ 1^{p^{t}_1}, 1^{p^{t}_2} \cdots 2^{p^{t}_1}, 2^{p^{t}_2} ... (N-1)^{p^{t-T}_K} \cdots N^{p^{t-T}_K} \}
\end{equation}

Here we define $P^{p^{t}_k}_n$ as the $k_{th}$ keypoint for the $n_{th}$ person in timestamp $t$.
The visualization of our proposed tokenization methods in demonstrated in Figure \ref{fig:token}.
After tokenizing the scene sequence as the four types of the aforementioned tokens, we linearly projected each token to four types of embedding metrics
and the output can be obtained by summing information of each type of token. That is $E = E_position + E_Type + E_Segment + E_Instance$.
Finally, the Action Tagger Network takes the embedding $E$ as input to make the actor-level action recognition for each token.

\subsection{Action Tagger Network}
\label{subsec:het_net}

The goal of the Action Tagger Network is to learn the spatial-temporal correlation of each keypoints $P^t$ in scene sequence $D$ and make the prediction for given downstream tasks (e.g. action recognition and action localization)
To achieve this, similar to make the prediction in sentence-level and token-level classification sub-task in BERT,
we feed embedding vector $E$ to a series of self-attention blocks to model the higher-order-interaction for every keypoints embedding vectors.
Then, we feed this representation to a fully-connected layers which is a learnable linear projection to make either sentence-level or token-level predictions.

\vspace{2mm}
\textbf{Transformer}
In our implementation of Transformers, we use the Transformers create three vectors from each of the input vectors (in our case the embedding of each keypoints). Hence, for each of the keypoint embedding, we create a projection for the Query vector (Q), a Key vector (K), and a Value vector (V). Next, we score for every keypoints of scene sequence $S$ against other keypoints by taking the dot product of the query vector(Q) with the key vector(K) the respective keypoints. Finally, it normalizes the score by $\sqrt{D}$ and a softmax operation. By multiplying each value vector(V) by the softmax score, the result can be obtained by summing up the weighted value vectors. The self-attention equation is as follow:

   \[ Attention(Q, K, V) = softmax(\frac{QK^T}{\sqrt{D}}) \]

\vspace{2mm}
\textbf{Hierarchical Transformer Encoder}
In our experiments, we find that as the length of input sequence increases, the computation complexity slows down the learning efficiency for the transformer due to its quadratic processing time, i.e., $O(n^2)$ for a sequence with n elements.
Hence, to address this quadratic inefficiency, for long sequences, instead of learning the self-attention weight of all keypoints in a single Transformer,
we replace it with our proposed Hierarchical Transformer Encoder that learn the action representation in a hierarchical manner.
Hence, given a keypoints embedding features $E_{\rho^t_n}$, a Keypoints Encode Transformer will first encode it into a list of action-level representation.
We follow \cite{Devlin2019} we take the first representation $h^{\rho^t_n}$ as the feature for an actor.

    \[E^{\rho^t_n} = (e_1^{\rho^t_n}, e_2^{\rho^t_n}, ... e_K^{\rho^t_n}) \]
    \[e_K^{\rho^t_n} = \rho^{p^{t}_k}_n + k^{p^{t}}_n \]
    \[h^{\rho^t_n} = Transformer(E^{\rho^t_n})\]

where $\rho^{p^{t}_k}_n$ is the Position Token and $k^{p^{t}}_n$ is the Type token.

Then, an Actor Encode Transformer will encode the actor-level representation $(h^{\rho^1_n}, h^{\rho^2_n} ... h^{\rho^{t-T}_N})$
to obtain context sensitive actor-level representations. $(d^{\rho_1}, d^{\rho_2}, ... d^{\rho_N})$
Finally, the actor level action classification is performed by linearly project $d^{\rho^n}$ to the number of total action classes in the given dataset.

    \[R^{\rho^t_n} = (r_1^{\rho^t_n}, r_2^{\rho^t_n}, ... r_K^{\rho^t_n}) \]
    \[r^{\rho^t_n} = h^{\rho^t_n} + P^{p^{t}_k}_n + T^{p_{k}}_n \]
    \[d^{\rho_n} = Transformer(R^{\rho_n})\]

where $P^{p^{t}_k}_n$ is the Instance Token and $T^{p_{k}}_n$ is the Segment Token.

\section{Experiments}
We evaluate the effectiveness of our approach on two tasks, action recognition, and action detection.
For action recognition, we report the performance on JHMDB and Kinetics datasets, reporting the Top-1 accuracy score.
For action localization, we report the performance on the AVA dataset, evaluating the mean average precision (mAP).
The content of this section is organized as follow:
First, we introduce the subsets of datasets used in our experiment in section \ref{exp:dataset}.
Then, we describe the implementation details in section \ref{exp:implementation}.
Finally, we report the performance of action recognition and action detection in \ref{subsec:exp:action_recognition} and \ref{subsec:exp:action_detection}, respectively.

\subsection{Dataset}
\label{exp:dataset}

\vspace{2mm}
\textbf{JHMDB Dataset\cite{Jhuang2013}.} JHMDB dataset is a pose-action recognition dataset that consists of 659 training videos and 267 testing videos.
It provides rich annotations including 15 joint positions, puppet mask, and puppet flow, which make it a good fit to evaluate  KeyNet utilizing the evolution of human joints as the major information to recognize human action.
In our experiments, we use this dataset as a starting point to validate if using only Keypoints as input modality is feasible for transformer-based architectures to recognize simple person movement actions.
For evaluation, we report the performance of action recognition in terms of accuracy on the first split of the JHMDB dataset.

\vspace{2mm}
\textbf{Kinetics-skeleton Dataset\cite{Yan2018}.} Kinetics-skeleton dataset is collected by providing extra annotation of human skeleton keypoints on the Kinetics\cite{Carreira2017} dataset.
Originally, the Kinetics dataset only provide coarse-grained action labels over the entire sequenece. Yan et al. \cite{Yan2018} use publicly available human pose estimator, Openpose\cite{Cao2021},
to extract 18 keypoints for the top two persons, in every scene, with the highest confidence scores, in terms of the summation of joint confidence scores.
In our experiments, we use this to validate if the proposed KeyNet can recognize action with keypoints annotation on different human body parts.
For evaluation, we manually select 16 action categories and report the performance in terms of accuracy.

\vspace{2mm}

\textbf{AVA Dataset\cite{Gu}.}
The Atomic visual Action (AVA) v2.1 consists of 211K, 57K, and 117K video clips for training, validation, and test sets.
The center frame or keyframe is taken at 1 FPS from 430 15-minute movie clips with dense actor level annotation
of all the bounding boxes and one or more among the 80 action classes.
For evaluation, our goal is to focus on validating the effectiveness and feasibility of keypoint based approach on multiple actors.
We sub-sample this dataset for two reasons. First, this dataset is heavily imbalanced, and even though RGB data can be augmented to handle class imbalance, improving class imbalances for pose information is rather difficult. Second, we identify the classes, where scene information provides the largest utility and test our methods specifically for those classes. 
Hence, to ease the high imbalance nature of AVA dataset, we manually select the 20 action classes that have more than 2000 samples including 8 classes of person movement actions (\textbf{P}),
4 classes of person-person interactive actions (\textbf{PP}) and 8 classes of person-object manipulation actions(\textbf{PO}).
For evaluation, we follow the official method of using frame-level mean average precision (frame-AP) at IOU threshold 0.5 as described in \cite{Gu}

\subsection{Experiment Details}
\label{exp:implementation}
% \textbf{Training}
In this subsection, we provide our experiment details, including  our hyperparameter settings and the data preprocessing procedure used in evaluating KeyNet.
We use Adam as the optimizer and design a learning rate schedule with a linear warmup.
The learning rate will warm up to the initial learning rate $\eta$ for a fraction of $0.01$ of total training iterations
and then linear decay to 0 as reaching the total training iteration.
For the action localization task on AVA dataset, we choose $N=5$ human tracklets and $M=3$ object masks to form the scene sequence and optimize our KeyNet model with batch size 32.
For the action recognition task, we choose $N=5$ human tracklets and $M=1$ object mask to form the scene sequence and optimize our KeyNet model with batch size 64

\vspace{2mm}
\textbf{Data Augmentation}
In our experiments, we found that data augmentation is a critical component to optimize the performance of our KeyNet.
Without the augmentation techniques, KeyNet tends to easily overfit on those majority classes. (e.g. stand, sit, talk to and watch actions in AVA dataset).
To solve this problem, we augment the training data with random flips, crops, and expand and further address the problem of the data imbalance with the $WeightedRandomSampler$ provided by Pytorch to equally sampled action categories in each training iteration before the estimation step.
As shown in Table \ref{tab:ava_aug}, adding data augmentation and re-sampling techniques can lead to a $+17.16\%$ performance gain in terms of mean average precision.

\begin{table}[t]
    \centering
    \small
    \begin{tabular}{c c c c c}
        \hline
        Action Type & Data Aug. & Weighted Sampler & mAP            \\
        \hline
        P           &           &                  & 14.25          \\
        P           & \cmark    &                  & 20.28          \\
        P           &           & \cmark           & 16.42          \\
        P           & \cmark    & \cmark           & \textbf{31.41} \\
        \hline
    \end{tabular}
    \caption{Ablation Study of techniques for the data imbalance in AVA dataset.}
    \label{tab:ava_aug}
\end{table}

\subsection{Performance on Action Recognition}
\label{subsec:exp:action_recognition}
Since recognizing action categories requires the awareness of both spatial and temporal domains, we first conduct experiments on small scale JHMDB dataset to determine
the best spatial-temporal configuration for our proposed KeyNet.
Then we generalize the task to kinetics dataset with more complex action and validate the effectiveness of using object keypoints to provide context features in videos.

\vspace{2mm}
\textbf{Spatial Resolution}
Spatial resolution is a key factor for recognizing human action on small scales.
Decreasing the spatial resolution caused the network to lose the fine-grained information but also reduce the computation cost.
To determine the trade-off between the recognition performance and the computation cost, we variate the resolution of Position Token and report the performance and the computation cost for KeyNet.
According to the statistic information in table \ref{tab:jhmdb_s_res}, the optimal resolution for position token is $32x24$.

\begin{table}[t]
    \centering
    \begin{tabular}{c c c c c}
        \hline
        Token Resolution & Accuracy       \\ [0.5ex]
        \hline
        32$*$24            & \textbf{55.81} \\
        64$*$48            & 53.55          \\
        96$*$72            & 50.41          \\
        128$*$96           & 37.99          \\
        \hline
    \end{tabular}
    \caption{Experiments for token resolution on the JHMDB dataset}
    \label{tab:jhmdb_s_res}
\end{table}

\vspace{2mm}
\textbf{Temporal Sequence length.}
Temporal  sequence length indicates the tokens along the temporal dimension which maps to the total number frames that the network processes from the input.
Especially for those actions with slow motion (e.g. taichi), it is necessary to increase the temporal sequence lengths to let our model fully capture the features of the entire action; however, this will cause the increases the computation.
In table \ref{tab:jhmdb_t_res}, we compare different configurations of temporal sequence lengths for our proposed KeyNet and find that the one with a longer temporal sequence tends to have worse performance indicating the difficulty of transformers in modelling longer sequences.
The longer sequence length prevents the self-attention layers in the  Transformer unit from learning the representative attention vectors for each type of action.
Therefore, in our following experiments, we fix the number of frames in our input as $10$ for a lower sequence length for the best performance.

\begin{table}[t]
    \centering
    \small
    \begin{tabular}{c c c c }
        \hline
        N Frames & Sequence Length & Token Size & Accuracy       \\ [0.5ex]
        \hline
        10       & 150             & 32$*$24      & \textbf{55.81} \\
        15       & 225             & 32$*$24      & 50.54          \\
        \hline
    \end{tabular}
    \caption{Experiments of temporal footprints on the JHMDB dataset}
    \label{tab:jhmdb_t_res}
\end{table}

\vspace{2mm}
\textbf{Effectiveness of Object Keypoints}
To demonstrate the effectiveness of our strategy that using object keypoints to compensate the context information,
we conduct experiments on JHMDB and Kinetics-16 dataset shown in Table. \ref{tab:object_kps}. We use the Kinetics-16 dataset  to evaluate object based action recognition while JHMDB is collected to evaluate the action of human body part motion or single-person action.
Our result show that the proposed methods improve the performance on the kinetics-16 dataset ($+4.5\%$) but also hurt the performance over the JHMDB dataset for a small margin ($-0.46\%$).
This occurs because JHMDB dataset has been designed for single person action, often with little to no correlation with objects in the scene. As a result for majority of the classes, this additional information, adds complexity to the input space and makes learning difficult.

\begin{table}[t]
    \centering
    \small
    \begin{tabular}{c c c }
        \hline
        Dataset     & Object Keypoints & Accuracy       \\
        \hline
        JHMDB       & \xmark           & \textbf{55.81} \\
        JHMDB       & \cmark           & 55.35          \\
        \hline
        % Kinetics-16 & CNN        & \xmark           & 44.00          \\
        % Kinetics-16 & CNN        & \cmark           & 47.20          \\
        Kinetics-16 & \xmark           & 45.40          \\
        Kinetics-16 & \cmark           & \textbf{49.90} \\
        \hline
    \end{tabular}
    \caption{Experiments of using object keypoints to provide context features in action recognition tasks}
    \label{tab:object_kps}
\end{table}

\subsection{Performance on Action Detection}
\label{subsec:exp:action_detection}
For this subsection, we describe the details about how to generalize our proposed methods to the action localization scenario to predict action for each actor in scene sequence $D$.
This is analogous to the correlation between sentiment analysis (sentence-level predictions) and Part of Speech Tagging (token-level predictions) in the Natural Language Process field.
The implementation can easily be done by replacing the last fully connected layer with a multi-label prediction layer.
However, we discover this poses two challenges: First, how to provide sufficient information to learn the complex interaction across each tracklet. Second, how to boost the learning efficiency of a long sequence of keypoints extracted from multi-person and multi-object data annotations.

\vspace{2mm}
\textbf{1 Frames Per Second.}
For the first challenge, the most intuitive way to provide additional information is to increase the temporal footprint.
As a result, we must address the issue of learning efficiency mentioned in  \ref{subsec:exp:action_recognition}.
Therefore, instead of collecting more frames in a scene sequence, we decrease the sampling rate in videos.
Our proposed workflow is described as followed:
First, we detect and estimate keypoints for human instances in all video frames. Then, we run a tracking algorithm for each of the detected bounding box starting from the key-frames.
Finally, we acquire the tracklets with different temporal footprints by sub-sampling the frames with specific intervals.
We report and analyze the performance of KeyNet with different temporal footprint settings to demonstrate the effectiveness of our proposed method.
As shown in in Table \ref{tab:ava_fps}, decreasing FPS from 5 to 1 can lead to $+3.24\%$ for person movement actions (P) and $+1.8\%$ for person movement and person-person interaction actions (PP).

\begin{table}[t]
    \centering
    \footnotesize
    \begin{tabular}{c  c c  c c}
        \hline
        Input Modality & FPS & Temporal Footprints & P     & P + PP \\
        \hline
        Keypoints      & 5   & 2s                  & 28.17 & 14.94  \\
        Keypoints      & 1   & 5s                  & 31.41 & 16.73  \\

        \hline
    \end{tabular}
    \caption{Comparison of temporal footprint and input modality. $P$ denotes the Person Movement actions and $PP$ denotes Person-Person interactive actions in terms of mean average precision (mAP)}
    \label{tab:ava_fps}

\end{table}

\vspace{2mm}
\textbf{Hierarchical Self-Attention Layer.}.
In table \ref{tab:ava_2stage_att}, we have shown that by learning the person-level and actor-level knowledge hierarchically,
our proposed hierarchical self-attention layer can improve the learning efficiency on the AVA dataset and lead to a $4.56 \%$ performance gain.
We also provide the performance of different configurations for transformer architecture.
According to table \ref{tab:ava:tx_arch}, the best configuration is using 6 heads, 4 hidden layers and layers with 128 hidden unit.
We follow this optimal setting for the following experiments.

\begin{table}[t]
    \centering
    \begin{tabular}{c c c}
        \hline
        Keypoints                   & Action Type & mAP   \\ [0.5ex]
        \hline
        Transformer                 &  P          & 26.85 \\
        Hierarchical Self-attention &  P          & 31.41 \\  [0.5ex]
        \hline
    \end{tabular}
    \caption{Effectiveness of our proposed hierarchical self-attention layer. Noted $P$ denotes the person-movement actions in the AVA dataset.}
    \label{tab:ava_2stage_att}
\end{table}

\begin{table}[t]
    \centering
    \footnotesize
    \begin{tabular}{c c c c c c}
        \hline
        N Heads & N Hidden & Hidden Size & Int.Size & Param. & mAP            \\
        2       & 4        & 128         & 128      & 0.91 M & 29.47          \\
        4       & 4        & 128         & 128      & 0.91 M & 29.45          \\
        6       & 4        & 128         & 128      & 1.78 M & \textbf{30.56} \\
        \hline
        4       & 4        & 64          & 256      & 1.99 M & 24.09          \\
        4       & 4        & 128         & 128      & 0.91 M & 29.45          \\
        4       & 6        & 128         & 128      & 5.97 M & 30.41          \\
        \hline
    \end{tabular}
    \caption{Experiments for the architecture searching for the proposed transformer-based architecture.}
    \label{tab:ava:tx_arch}
\end{table}

\vspace{2mm}
\textbf{Object Keypoints.}
We evaluate the effectiveness of object keypoints on all of the selected actions in the AVA dataset, including person movement(P), person-person interaction(PP), and person-object manipulation (PO) action categories.
Based on the statistical information in table \ref{tab:ava_object_kps}, the model with object keypoints to compensate context information loss has superior performance to the one without object keypoints.

\begin{table}[h]

    \centering
    \begin{tabular}{c c c}
        \hline
        Object Keypoints & Action Type & mAP   \\ [0.5ex]
        \hline
        \xmark           & P + PP + PO & 11.23 \\
        \cmark           & P + PP + PO & 11.45 \\  [0.5ex]
        \hline
    \end{tabular}
    \caption{Effectiveness of addressing object keypoints to provide contextual features.}
    \label{tab:ava_object_kps}
\end{table}

\begin{table}[b]
    \centering
    \begin{tabular}{c c c}
        \hline
        Input Modality & Action Type & mAP    \\
        \hline 
        RGB            & P           & 18.12   \\
        Keypoints      & P           & 31.41   \\
        \hline
        RGB            & P + PP      & 15.85   \\
        Keypoints      & P + PP      & 16.73   \\
        \hline
        RGB            & P + PP + PO &  9.28   \\
        Keypoints      & P + PP + PO & 11.45   \\
        \hline
    \end{tabular}
    \caption{The demonstration of context information recovery by comparing the performance of using full images and keypoints as input modality. $P$ denotes the Person Movement actions and $PP$ denotes Person-Person interactive actions in terms of mean average precision (mAP)}
    \label{tab:ava_kps_rgb}

\end{table}

\vspace{2mm}
\textbf{Context Information Recovery}
To demonstrate the effectiveness of using the object keypoints to recover context information in videos, we design a transformer-based RGB baseline to compare the recognition performance with Keypoints-based methods.
For the RGB baseline, we directly takes the image-level features from HRNet\cite{Wang2019} and Mask-RCNN\cite{He2020} as actor and context features. Then we feed the actor and context features to the same Action Tagger Network with our KeyNet.
In Table \ref{tab:ava_kps_rgb}, it clearly shows that KeyNet, using only keypoints, can achieve better performance than the RGB-baseline and based on our results, we believe that using human and object keypoints as a structure representation has the potential to fully recover the essential context information for action recognition.

\section{Conclusion}
In this work, we demonstrate that using the object-based keypoints informatio can compensate for accuracy loss due to the missing context information in keypoint-based methods. We also show a method to extract object keypoints from segmentation information and build a structure representation with human keypoints from videos. According to our experimental results, we have validated our proposed KeyNet has superior performance to the RGB baseline, a method based on image-level information, and shows the potential of using only keypoints to recover essential context information for action recognition.

% \section{Acknowledgements}
% Deep Patel did not contribute to IP development. His core responsiblities were towards data curation, running experiments and hyperparameter search. 

%%%%%%%%% END BODY TEXT
\balance
{\small
    \bibliographystyle{ieee_fullname}
    \bibliography{egbib}
}

\end{document}